\documentclass[letterpaper, 10 pt, conference]{ieeeconf}  

\IEEEoverridecommandlockouts                              

\usepackage{epsfig} 
\usepackage{calc}
\usepackage{xcolor}

\usepackage{array}

\usepackage{amsmath,amsfonts,amssymb} 

\usepackage{float}

\usepackage[linesnumbered,ruled]{algorithm2e}

\usepackage{pgfplots}

\usepackage{graphicx}
\usepackage{subcaption}
\usepackage{todonotes}
\usepackage{changepage}

\ifdefined\r
\renewcommand{\r}[3]{
    \mathbf{r}^{#1}_{#2/#3}
}
\else
\newcommand{\r}[3]{
    \mathbf{r}^{#1}_{#2/#3}
}
\fi

\newlength{\mylengthL}
\newlength{\mylengthR}

\title{\LARGE \bf
Learning Continuous Occupancy Maps with the Ising Process Model
}

\author{Nicholas O'Dell, Christopher Renton, and Adrian Wills
\thanks{}
\thanks{
  Nicholas O'Dell, Christopher Renton and Adrian Wills are with the School of Engineering, University of Newcastle Australia, NSW, 2308,{\{\tt\small Nicholas.Odell, Christopher.Renton, Adrian.Wills\}@newcastle.edu.au}
}%
}

\begin{document}

\maketitle
\thispagestyle{empty}
\pagestyle{empty}

\begin{abstract}

We present a new method of learning a continuous occupancy field for use in robot navigation. Occupancy grid maps, or variants of, are possibly the most widely used and accepted method of building a map of a robot's environment. Various methods have been developed to learn continuous occupancy maps and have successfully resolved many of the shortcomings of grid mapping, namely, priori discretisation and spatial correlation. However, most methods for producing a continuous occupancy field remain computationally expensive or heuristic in nature. Our method explores a generalisation of the so-called Ising model as a suitable candidate for modelling an occupancy field. We also present a unique kernel for use within our method that models range measurements. The method is quite attractive as it requires only a small number of hyperparameters to be trained, and is computationally efficient. The small number of hyperparameters can be quickly learned by maximising a pseudo likelihood. The technique is demonstrated on both a small simulated indoor environment with known ground truth as well as large indoor and outdoor areas, using two common real data sets.
\end{abstract}

\section{INTRODUCTION} 
\label{sec:introduction}

Robots are increasingly operating in environments where they are interacting with their surroundings. Therefore, the ability of a robot to map the surrounding environment is one of the major requirements for it to be considered truly autonomous \cite{thrun2005probabilistic}. Mapping is an infinite dimensional problem, as there are infinitely many ways an environment can be populated, this is what makes mapping challenging. 

There are two main types of maps. Feature maps detect and store a finite list of landmarks and their spatial locations. Occupancy grid maps represent a robot's environment as a uniform grid of cells, each of which is \emph{either} occupied or free space. Maps of this nature overcome many of the challenges of feature based maps and make fewer assumptions about the environment of the robot, however, the world is continuous in nature, therefore any occupancy grid is a high dimensional approximation of an infinite dimensional occupancy map. Several modelling assumptions have led to both the computational efficiency and, in many situations, inaccuracy of the grid map. 

One such assumption is that each grid cell is independent of the others. This leads to an unintuitive result; decreasing the cell size (increasing resolution) can lead to a decrease in information about the map with the same measurement set. Furthermore, as the grid size diminishes, the probability of intersecting the same grid cell also goes to zero (meaning that re-observing the same cell and therefore localisation are impossible). Another major consideration when choosing the cell size is the size of the space which is to be mapped, which may put a lower bound on the cell size due to storage requirements. This is difficult in online applications when the size of space being explored is generally unknown beforehand. However, even with its limitations the grid map and its variants remain one of the most useful, widely used, and feasible to implement solutions available for mapping a robot's environment.

Since the development of the grid map there has been several attempts to relax the independence assumptions by solving the occupancy mapping problem in a continuous space. 
The method presented in \cite{thrun2003learning} introduced dependencies between grid cells and successfully achieved more accurate occupancy grids when using wide angled sensors, in situations where traditional grid mapping performs poorly. However, this method still uses the occupancy grid, which is a discretisation of an occupancy field. More recent developments formulate the occupancy mapping problem in its original infinite dimensional space, generally referred to as continuous occupancy mapping techniques. 

Gaussian process occupancy mapping (GPOM) \cite{o2009contextual} represents the first attempt to learn a continuous occupancy field using kernel methods, specifically, Gaussian processes (GP). This method exploits the fact that environments naturally contain structure to model context in the robot's surroundings with a GP. By modelling structure in the environment, accurate inference can be achieved in partially observed and occluded areas. Grid maps typically perform poorly in this situation due to the independence assumption. Hilbert maps were introduced as a less computationally expensive alternative to GPOM \cite{ramos2016hilbert}. The concept is to adopt a logistic regression classifier operating on a high dimensional feature vector generated from efficient kernel approximations. The classifier is then trained with fast stochastic gradient descent.
A more recent development places a beta distribution on the expected value of occupancy and updates its parameters recursively \cite{doherty2017bayesian}. This method utilises non-parametric Bayesian inference techniques for exponential families \cite{vega2014nonparametric}.

With a few exceptions, kernel methods that have been adapted for mapping so far are either computationally infeasible for online applications or are heuristic in nature. GPOM is not only computationally expensive, $\mathcal{O}(n^3)$, but also restrictive as the measurement models are limited to linear mappings \cite{hennig2013quasi}. Hilbert mapping is over parametrised requiring thousands of parameters to be trained online.




 In this paper we propose a novel alternative based upon an extension of the Ising model, where each binary random variable is indexed by a spatial location, defining a stochastic process. By adapting common kernel functions used in Gaussian process regression to obtain spatial context, we are able to complete exact inference using all of the available measurements. 



The contributions of this paper are:
\begin{enumerate}
  \item The proposed mapping technique, a probabilistic method of performing exact inference on a binary occupancy field.
  \item A novel range sensor model for use within our method.
  \item A means of training the required hyperparameters using the so-called pseudo likelihood.
\end{enumerate}


\section{OCCUPANCY MAPPING} 
\label{sec:occupancy_mapping}

Prior to the invention of the occupancy grid map, parametric maps such as landmark maps dominated the field of robot perception and navigation. The feature mapping paradigm requires a database of pre-defined features for mapping. This may be unsuitable in environments where features are sparse or the environment does not contain any expected features.

Although grid maps do not pre-suppose the contents of a robot's environment, they do make assumptions about an environment's geometric structure and assumptions about spatial correlation. Grid maps have the limitation of assuming that a lattice process, with a finite number of cells, is sufficient to approximate an occupancy field, which is an infinite dimensional object. Furthermore, grid maps assume that the entirety of each cell is either occupied or free, which leads to unsatisfactory results when cells are partially occupied. All physical locations within any given cell have the same probability of occupancy (perfectly correlated) while the occupancy of locations outside a cell, even those very close to the cell border are completely uncorrelated.

\begin{figure}[h]
\vspace{2mm}
\centering
\begin{minipage}{0.5\linewidth}
  \centering
\includegraphics[width = \textwidth]{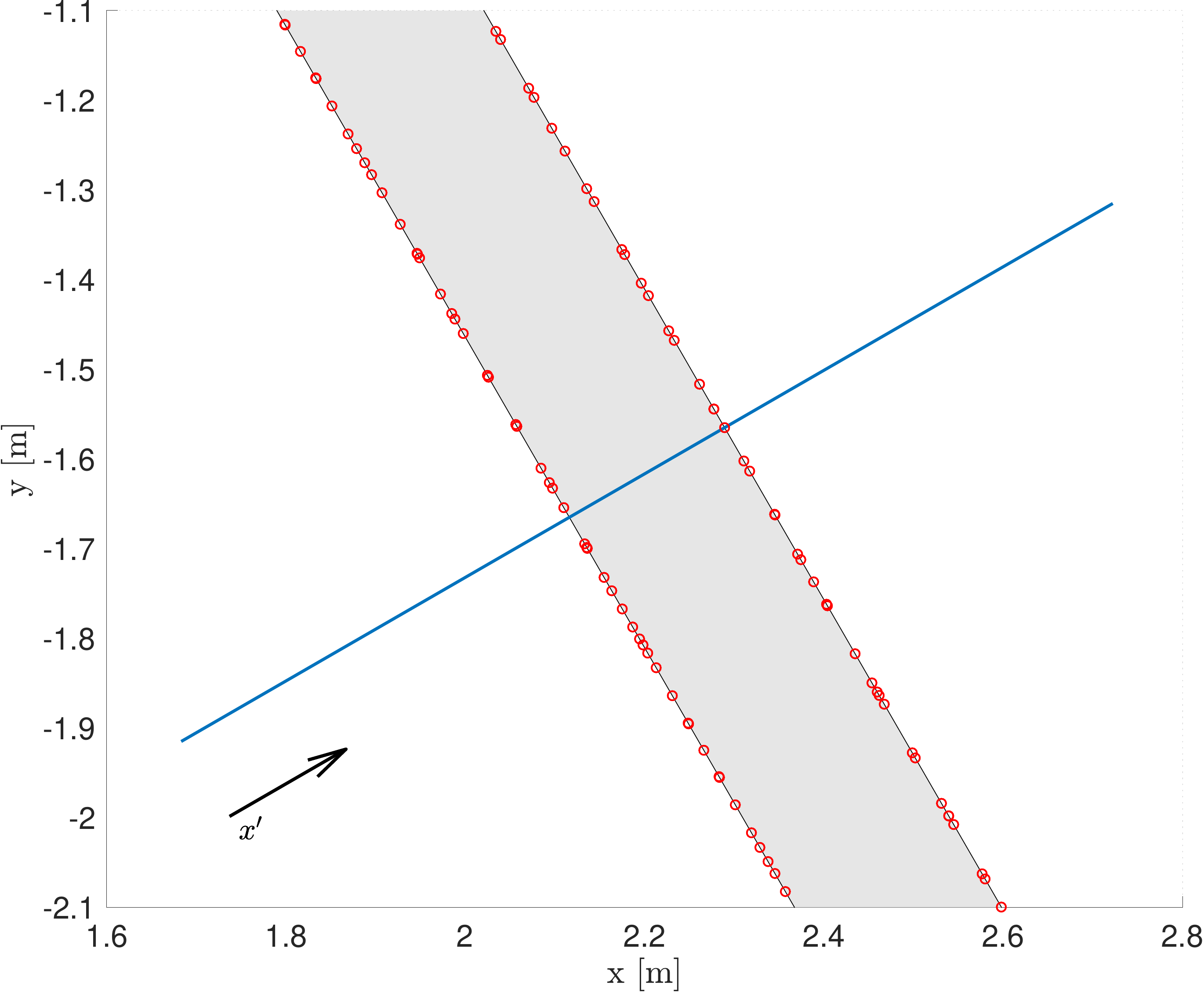}
\label{fig:truth_wall}
\end{minipage}%
\hfill
\begin{minipage}{0.5\linewidth}
  \centering
\includegraphics[width = \textwidth]{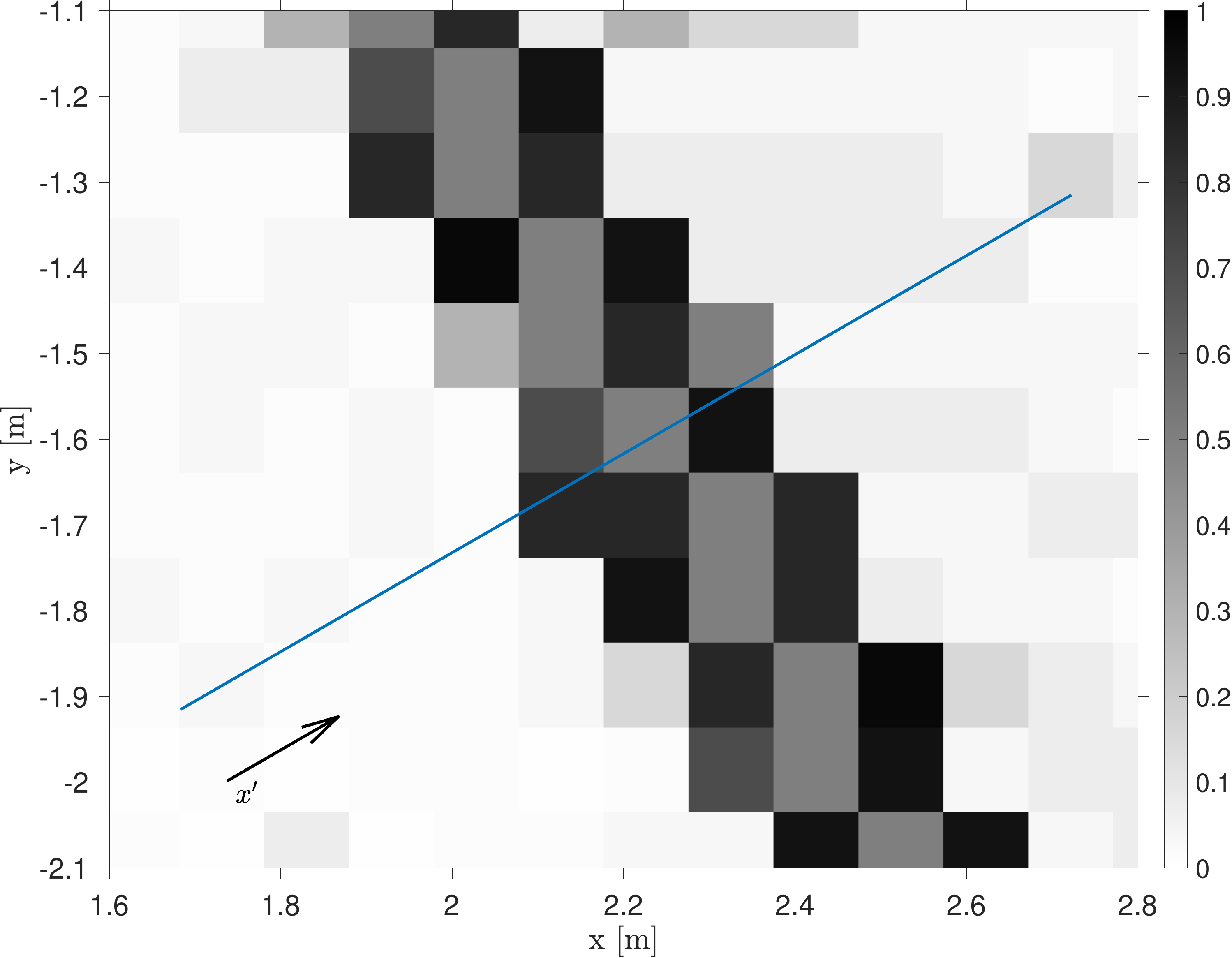}
\label{fig:grid_wall}
\end{minipage}\\
\begin{minipage}{0.5\linewidth}
  \centering
\includegraphics[width = \textwidth]{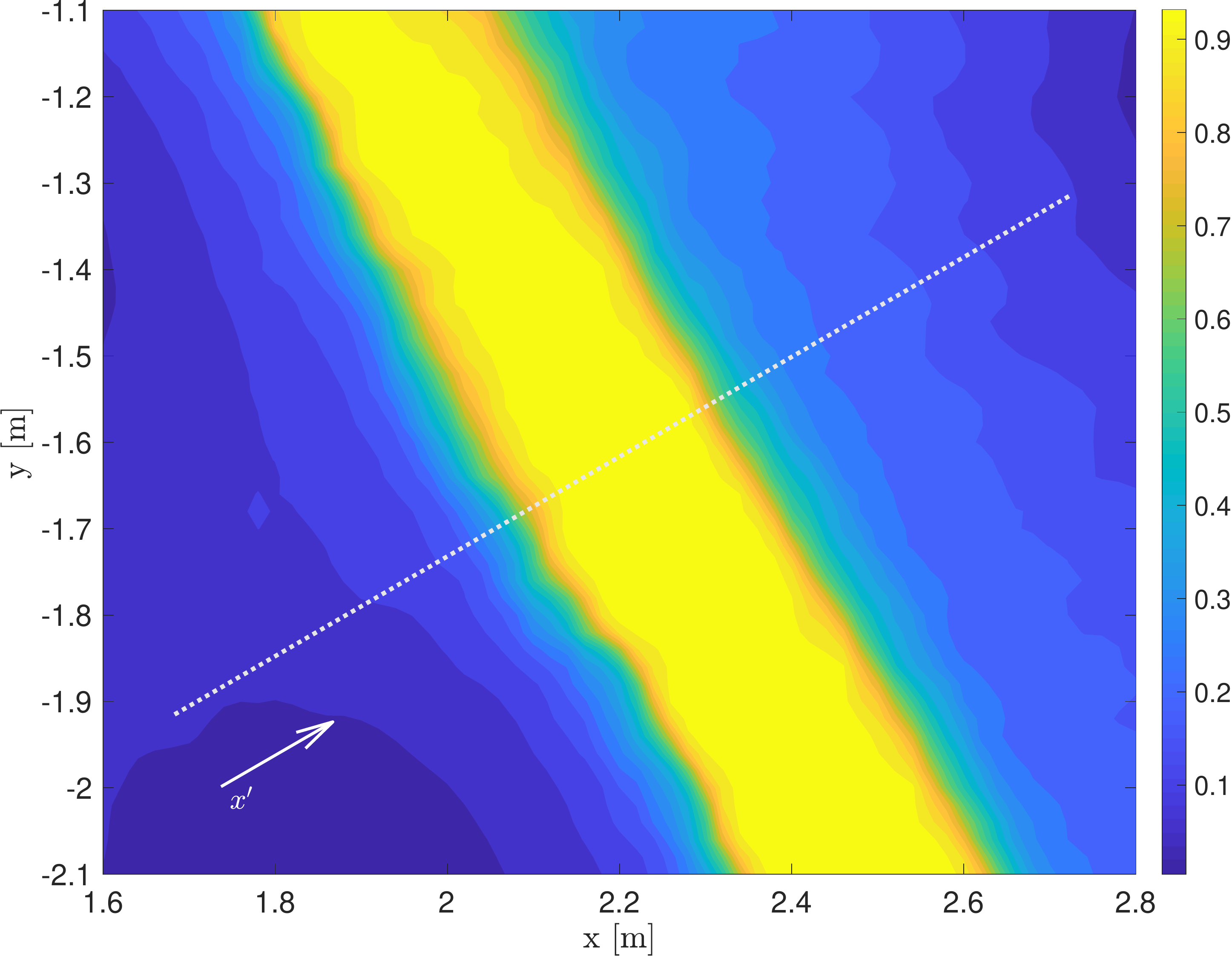}
\label{fig:ipom_wall}
\end{minipage}%
\hfill
\begin{minipage}{0.5\linewidth}
  \centering
\includegraphics[width = \textwidth]{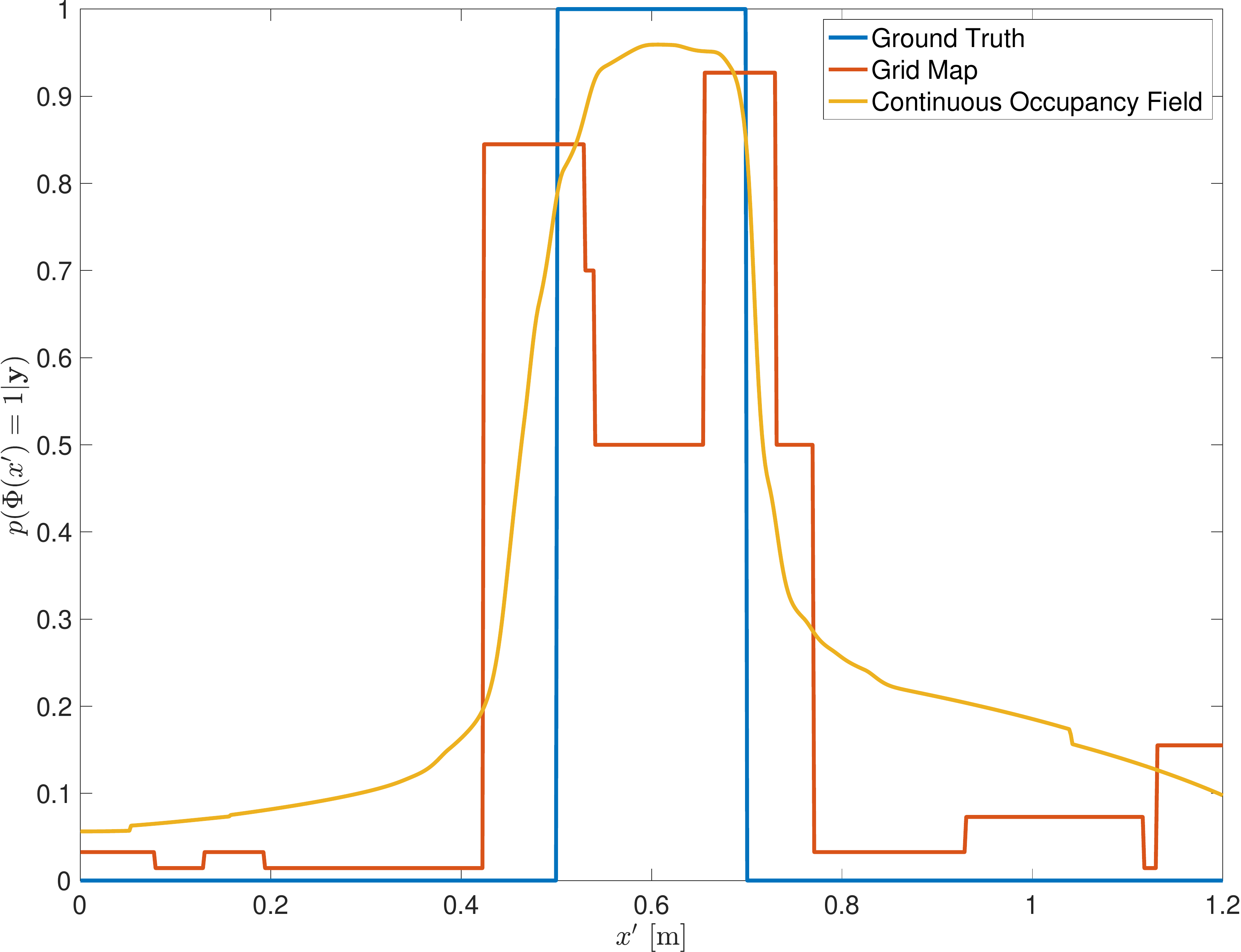}
\label{fig:compare_wall}
\end{minipage}%
\caption{(a) shows the cross section of a wall and hit points from a lidar. (b) and (c) shows the maps learnt by grid mapping and our proposed method respectively, compared with the ground truth (d). }

\label{fig:smallExample}
\end{figure}

A canonical example of the grid map's assumptions having a detrimental effect on the results is shown in Fig.~\ref{fig:smallExample}. 
Because of the geometric structure of fixed-cell-size grid maps there is no opportunity to improve the grid maps geometry over time. The example depicts the probability of occupancy through a cross section of a wall, showing from left to right as the robot obtains more measurements on both sides of the wall. Because the grid map has a fixed geometry it is unable to learn the true geometry of the map as more measurements are obtained, instead these measurements provide conflicting information and produce an inaccurate result. Because our proposed method learns a continuous occupancy field instead of an occupancy grid we are able to more accurately learn the structure of the environment as more measurements are obtained, gaining the same advantages as other continuous occupancy mapping methods, with a similar run-time as a grid map.

%


\section{\textcolor{black}{Proposed mapping method}} 
\label{sec:ising_process_occupancy_mapping}

The Ising model was developed in 1925 for modelling ferro-magnetism in statistical mechanics \cite{niss2005history}. A field of binary random variables are placed in a graph structure, usually in a lattice pattern, and represent the magnetic dipole moments of atomic spins. The model allows each of the random variables to interact with its nearby neighbours. 

The joint distribution for a set of binary random variables with first and second order coupling information which maximises entropy takes the form of an Ising model with pairwise interactions \cite{broderick2007faster}. 
Maximum entropy models are being used increasingly in several fields, including neuroscience  and psychometrics \cite{albanna2012minimum,epskamp2016network}. 

The most general form of the Ising model is

\begin{equation}\label{eq:ising}
    p(\mathbf{s})=\frac{1}{\mathcal{Z}_\mathbf{s}} \exp\left( \mathbf{s}^\mathsf{T} \boldsymbol\lambda  + \frac{1}{2}\mathbf{s}^\mathsf{T} \boldsymbol\Lambda \mathbf{s} \right),
\end{equation}
where each binary random variable $s_i\in\{-1,1\} \ \forall i$, $\boldsymbol\Lambda$ defines second order coupling information,  $\boldsymbol\lambda$ describes first order statistics and $\mathcal{Z_\mathbf{s}}$ is the normalising constant. 


\subsection{Map Inference} 
\label{sub:ising_map_inference}
As the Ising model is a distribution which represents spatially correlated binary random variables in a field, it appears to be a suitable candidate to model an occupancy field. 
For the purposes of occupancy mapping we will assume that the robot pose, $(\r{n}{B}{N}, \mathbf{R}^n_b)$ is known. Where $\r{n}{B}{N}$ is the vector sensor's position $(B)$ with respect to the origin $(N)$ expressed in the inertial coordinate system $\{n\}$ and $\mathbf{R}^n_b$ is the transformation between the body fixed coordinate system $\{b\}$  and the inertial coordinate system $\{n\}$ systems. Additionally, we will only be focussing on range measurements obtained from lidar type sensors, which for the purpose of mapping we will model as a unit direction vector $\r{b}{Q_i}{B}$ and range $d_i$. Each measurement $y_i$ is therefore
\begin{equation}
y_i = \{ d_i, \mathbf{r}^b_{Q_i/B},\mathbf{r}^n_{B/N}, \mathbf{R}^n_b \}.
\end{equation}
\subsubsection{Ising Map Update} 
\label{ssub:ising_map_update}
Firstly let us introduce suitable notation: The scalar function $\Phi \colon \mathbb{R}^d \to \{ -1,1\}$, with $d \in\{2,3\}$ depending on the dimension of the map. $\Phi(\mathbf{x})$ is a binary random variable representing the occupancy of $\mathbf{x}$, where

\begin{equation}
\Phi(\mathbf{x})=\left\{\begin{array}{ll}{+1} & { : \mathbf{x} \in \mathbb{O} \ (\text {Occupied})} \\ {-1} & { : \mathbf{x} \in \mathbb{F} \ (\text {Free})}\end{array}\right\}.
\end{equation}

Similarly, we introduce the shorthand

\begin{equation}
\boldsymbol\Phi^\mathbf{x} = \begin{bmatrix}
  \Phi(\mathbf{x}_\mathbf{1})\\
  \vdots\\
  \Phi(\mathbf{x}_\mathbf{n})
\end{bmatrix}
,
\end{equation}
for a vector of binary r.v.s modelling occupancy of points $\mathbf{x_1}, \ldots, \mathbf{x_n}$. We propose that the prior occupancy of these points be distributed according to
\begin{equation} \label{eq:isingprior}
p(\boldsymbol\Phi^\mathbf{x}) = \frac{1}{\mathcal{Z}} \exp \left( \left(\boldsymbol\Phi^\mathbf{x}\right)^{\mathsf{T}} \boldsymbol\lambda^\mathbf{x}_{y_0}  + \frac{1}{2} \left(\boldsymbol\Phi^\mathbf{x}\right)^{\mathsf{T}} \boldsymbol{\Lambda}^{\mathbf{x}}_{y_0} \boldsymbol\Phi^\mathbf{x}\right),
\end{equation}
where $\boldsymbol\lambda^{\mathbf{x}}_{y_0}$ and $\boldsymbol\Lambda^{\mathbf{xx}}_{y_0}$ are the first and second order statistics of the prior. The superscript represents the spatial locations and the subscript indicating which measurements the distribution has been conditioned on, $y_0$ indicating no measurements have been incorporated.

We would like to enforce closure such that a posterior distribution is of the same form as the prior, enabling us to set up a recursive solution for $\boldsymbol\lambda^{\mathbf{x}}$ and $\boldsymbol\Lambda^{\mathbf{xx}}$. 
We seek a likelihood model such that the prior is conjugate. 
If we use
\begin{equation}\label{eq:likelihood}
p(y \vert \boldsymbol\Phi^\mathbf{x} ) \propto \exp \left( \left(\boldsymbol\Phi^\mathbf{x}\right)^{\mathsf{T}} \boldsymbol\lambda(y,\mathbf{x})  + \frac{1}{2} \left(\boldsymbol\Phi^\mathbf{x}\right)^{\mathsf{T}} \boldsymbol{\Lambda}(y,\mathbf{x}) \boldsymbol\Phi^\mathbf{x}\right)
\end{equation}
as a likelihood function, after utilising Bayes' theorem


\begin{equation}
\begin{split}
  p(\boldsymbol\Phi^\mathbf{x} \vert \mathbf{y}_{1:n}) \propto \exp \left( \left(\boldsymbol\Phi^\mathbf{x}\right)^{\mathsf{T}} \boldsymbol\lambda^\mathbf{x}_{\mathbf{y}_{1:n}}  + \frac{1}{2} \left(\boldsymbol\Phi^\mathbf{x}\right)^{\mathsf{T}} \boldsymbol{\Lambda}^{\mathbf{x}}_{\mathbf{y}_{1:n}} \boldsymbol\Phi^\mathbf{x}\right),
\end{split}
\end{equation}
where the recursion for $\boldsymbol\lambda^\mathbf{x}$ and $\boldsymbol\Lambda^\mathbf{x}$ is given by
\begin{equation}
\begin{split}
  \boldsymbol\lambda^\mathbf{x}_{\mathbf{y}_{1:n}} &= \boldsymbol\lambda^\mathbf{x}_{\mathbf{y}_{1:n-1}} + \boldsymbol\lambda(y_n ,\mathbf{x}), \\
  \boldsymbol\Lambda^\mathbf{x}_{\mathbf{y}_{1:n}} &= \boldsymbol\Lambda^\mathbf{x}_{\mathbf{y}_{1:n-1}} + \boldsymbol\Lambda(y_n,\mathbf{x}).\\
\end{split}
\end{equation}

In this work we have neglected the second order statistic $\boldsymbol\Lambda^{\mathbf{x}}$ for several reasons. Firstly, to generate and plot a map of a reasonable resolution would mean marginalising over somewhere in the order of $10^{300}$ combinations. Neglecting this term instead leads to a very tractable and fast algorithm. Secondly, neglecting this term creates an independence assumption, the effect of which is decreased by modelling spatial correlation in the likelihood function, shown in the following section. With these assumptions the posterior probability of occupancy is now
\begin{equation}
  p(\boldsymbol\Phi^\mathbf{x} \vert \mathbf{y}_{1:n}) \propto \exp \left( \left(\boldsymbol\Phi^\mathbf{x}\right)^{\mathsf{T}} \boldsymbol\lambda^\mathbf{x}_{\mathbf{y}_{1:n}}  \right),
\end{equation}
and the posterior probability of each variable can be calculated independently using
\begin{equation}
\ell\left( \Phi(\mathbf{x}_i) \right) = 2\lambda_{\mathbf{y}_{1_n}}^{\mathbf{x}_i},
\end{equation}
where $\ell\left( \Phi(\mathbf{x}_i) \right)$ is the log odds probability of $\mathbf{x}_i$ being occupied and $\lambda_{\mathbf{y}_{1:n}}^{\mathbf{x}_i}$ is the statistic associated with location $\mathbf{x}_i$.




\subsubsection{Sensor Model} 
\label{ssub:sensor_model}
To utilise this method for mapping, an appropriate sensor model needs to be employed to determine the parameters of \eqref{eq:likelihood}. 
Several methods have been explored for modelling range measurements within the occupancy mapping setting. Such as sampling the beam at regular intervals or modelling the beam as a line integral of free space along the beam with a point measurement of occupancy at the termination point \cite{o2011continuous}. We are less restricted in our setting as there are no constraints on the parameters of \eqref{eq:likelihood}.

\begin{figure}[h]
    \centering
    \begin{tikzpicture}
    [scale=3]
    \coordinate (B) at (0.1,0.5);
    \node (N) at (0,0) [below]{N};
    \coordinate (R) at (1.5,0.9);
    \coordinate (P) at (0.8,0.5);
    \coordinate (proj) at (0.74,0.681);

    \draw[->,very thick] (N) -- (B) node[midway, left]  {$\mathbf{r}_{B/N}$};
    \draw[->,very thick] (B) -- (R) node[midway, above] {$\mathbf{r}_{H/B}$};
    \draw[->,very thick] (B) -- (P) node[midway, below] {$\mathbf{r}_{P/B}$};
    \draw[->,very thick,red] (P) -- (proj);
    \draw[->,very thick,blue] (B) -- (proj)node[midway, above] {$\text{proj} _{\mathbf{r}_{\text{\scalebox{0.5}{$H/B$}}}}$\hspace{-0.1cm}$\text{\scalebox{1.2}{$\mathbf{r}_{\text{\scalebox{0.7}{$P/B$}}}$}} \qquad$};
    \node(p) at (P) [right]{$P$};
    \node(b) at (B) [above]{$B$};
    \node(r) at (R) [right]{$H$};
\end{tikzpicture}
\caption{Vector diagram depicting the sensor $B$ obtaining a hit from $H$.}
\label{fig:vecfig}
\end{figure}
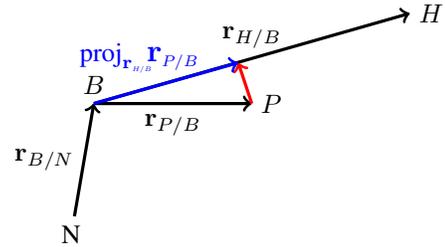

Fig.~\ref{fig:vecfig} shows the position of a range sensor ($B$), and the termination point of a beam ($H$), a point where occupancy is being inferred ($P$), $\mathbf{r}_{B/N}$ is the pose of the robot and $N$ is the origin of the map coordinates.
Information is obtained about a free space \emph{measurement} along the thin beam, a \textit{measurement} of occupancy at $H$ and no knowledge is gained about beyond the termination point. 

\begin{figure}[h]
\centering
  \centering
\includegraphics[width = 0.9\linewidth]{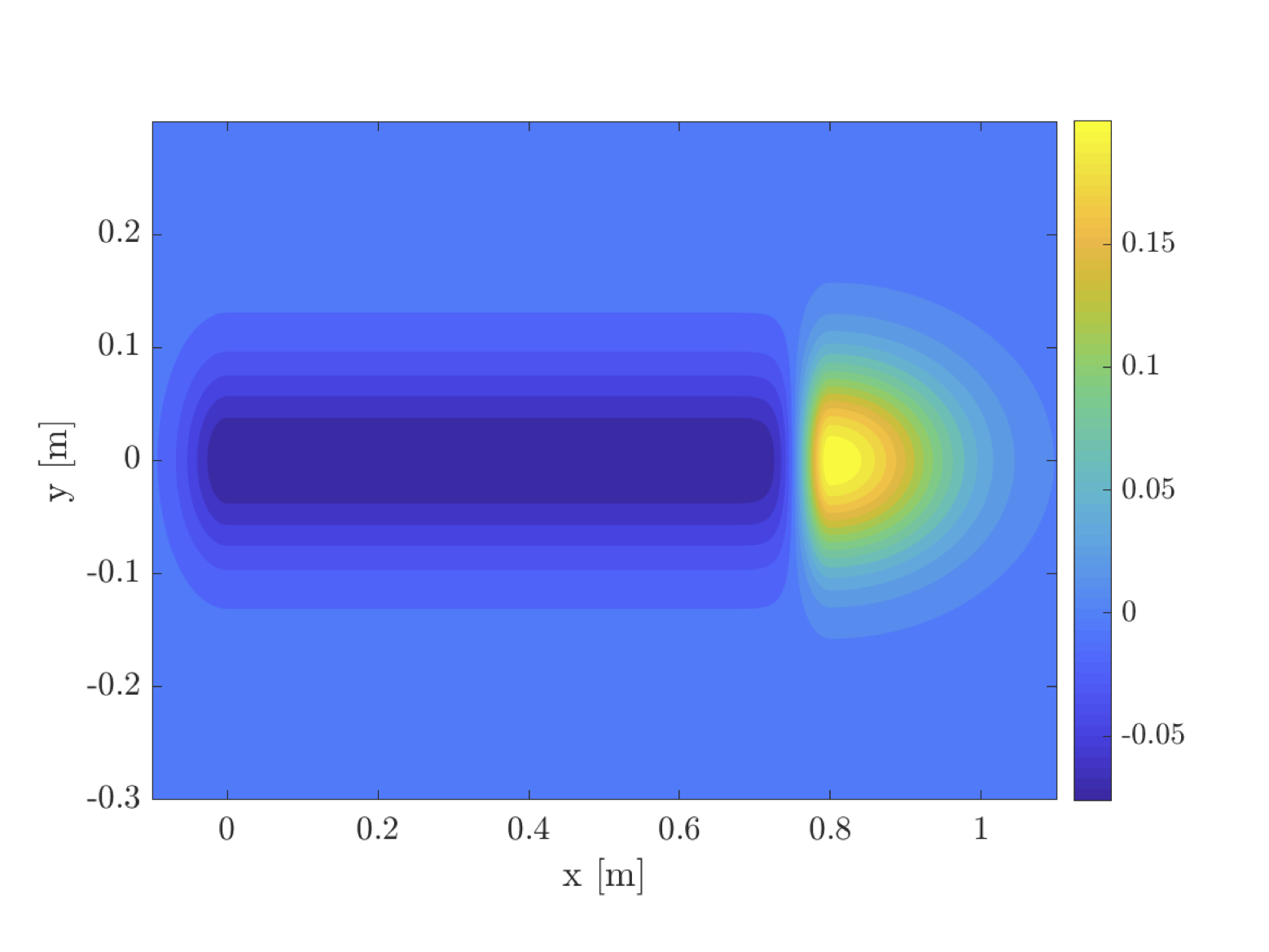}
\caption{Likelihood function output for $y_i = \{ d_i = 0.8, \mathbf{r}^b_{Q/B} = [1 \ 0]^{\mathsf{T}},\mathbf{r}^n_{B/N} = [0 \ 0]^{\mathsf{T}}, \mathbf{R}^n_b = \mathbf{I}_2\}$ at various locations $\mathbf{x} = [x,y]^\mathsf{T}$.}\label{fig:hitkern}
\label{fig:kerns}
\end{figure}

Alg.~\ref{alg:hit} encodes a positive correlation that decays radially about the point that a hit was returned from, but drops off fairly quickly in-front of the hit. Additionally, between the sensor and the hit, negative correlation decays to zero as a function of the perpendicular distance from the point $P$ to the beam. The result of this function is shown in Fig.~\ref{fig:hitkern} for an example measurement.
\IncMargin{0.5em}
\begin{algorithm}
    \SetKwInOut{Input}{Input}
    \SetKwInOut{Output}{Output}

    \Input{$\mathbf{r}^n_{B/N}$(pose), $\mathbf{R}^n_b$ (rotation matrix), $\mathbf{r}^b_{H/B}$ (hit), $\mathbf{r}^n_{P/N}$ (query)}
    $\r{n}{H}{B} = \mathbf{R}^n_b \ \r{b}{H}{B}$\\
    $\r{n}{P}{B} = \r{n}{P}{B} - \r{n}{B}{N}$\\
    $\mathbf{M} = (\r{n}{H}{B} \cdot \r{n}{P}{B})/(\r{n}{H}{B} \cdot \r{n}{H}{B})$\\
    $\mathbf{v}_1 = M \times \r{n}{H}{B}$\\
    $\mathbf{v}_2 = \r{n}{P}{B} - \mathbf{v}_1$\\
    $\mathbf{v}_3 = \r{n}{H}{B} - \mathbf{v}_1$\\
    \uIf{$M\geq 0 \boldsymbol\wedge M<1$}{
    $\mathcal{K} = (\sigma_h + \sigma_f) \exp (-0.5 \mathbf{v}_3 \cdot \mathbf{v}_3/l_f^2) - \sigma_f$\\
    $\lambda = \mathcal{K} \exp (-0.5 \mathbf{v}_2 \cdot \mathbf{v}_2/l_p^2)$\\
    }
    \uElseIf{$M\geq 1$}{
    $\mathcal{K} = (\sigma_h) \exp (-0.5 \mathbf{v}_3 \cdot \mathbf{v}_3/l_b^2) $\\
    $\lambda = \mathcal{K} \exp (-0.5 \mathbf{v}_2 \cdot \mathbf{v}_2/l_p^2)$\\
    }
    \Else{
    $\mathcal{K} = (\sigma_f) \exp (-0.5 \mathbf{v}_1 \cdot \mathbf{v}_1/l_f^2) $\\
    $\lambda = \mathcal{K} \exp (-0.5 \mathbf{v}_2 \cdot \mathbf{v}_2/l_p^2)$\\
    }

      \Return{$\lambda$}
    \caption{Likelihood Model - Hit ($\lambda(y_i,\mathbf{x})$)}
    \label{alg:hit}
\end{algorithm}
\DecMargin{0.5em}



\subsection{Hyperparameter Selection} 
\label{sub:hyper_parameter_selection}
Hyperparameter selection is an important aspect of model selection. We have several levels of design choices. At a higher level we may choose the class of likelihood functions we consider, while 
at a lower level we must choose the hyperparameters which govern the inference. In our case we must choose the set of hyperparameters used in Alg.~\ref{alg:hit}, defined as
\begin{equation}
\boldsymbol\theta = \{ \sigma_f, \sigma_h, l_p, l_f, l_b \}.
\end{equation}

Hyperparameter optimisation is often done by maximising a suitable likelihood function. The log-marginal-likelihood of the data is typically chosen as a suitable candidate, and has a long history of use in spatial statistics \cite{mardia1984maximum}.
The log-marginal-likelihood has a built in trade off between model fit and model complexity \cite{rasmussen2006gaussian}. Unfortunately in the case of the Ising model, the marginals and their gradients are intractable due to the partition function. Rather than approximate the true marginal likelihood we employ a pseudo-likelihood\footnote{Also called leave one out cross validation (LOO-CV).} as a suitable cost function.

The use of a pseudo likelihood was proposed by \cite{besag1974spatial} as a tractable alternative to the true likelihood.
 The pseudo likelihood  approximates the true marginal likelihood as a product sum of individual conditional probabilities \cite{besag1974spatial}. The pseudo likelihood is generally defined as

  \begin{equation}
    p(\mathbf{y}_{1:n} ; \boldsymbol\theta)= \prod _i^n p(y_i|\mathbf{y}_{\neg i};\boldsymbol{\theta}),\\
    \end{equation}
    or
    \begin{equation}
     \log p(\mathbf{y}_{1:n} ; \boldsymbol\theta)= \sum _i^n \log p(y_i|\mathbf{y}_{\neg i};\boldsymbol{\theta}),
    \end{equation}
where 

\begin{equation}
\mathbf{y}_{\neg i} = \{ y_j \mid j\neq i \}.
\end{equation}


The optimisation was carried out by maximising 
  \begin{equation}\label{eq:logPLIPOM}
\begin{split}
    \log p(\mathbf{y}_{1:n}; \boldsymbol\theta) &= \sum_i^n \log p(\phi_i|\mathbf{y}_{\neg i};\boldsymbol{\theta}) \\
    & + \sum_i^n \log p(\psi_i|\mathbf{y}_{\neg i};\boldsymbol{\theta}),
\end{split}
    \end{equation}
 where the end point of the $i$-th beam was considered to be a measurement of occupied space, $\phi_i = 1$. Additionally a randomly selected point along each beam was considered to be a pseudo measurement of free space, $\psi_i = -1$. For every $\phi_i$ and $\psi_i$ the dependence on the corresponding $y_i$ was removed. A similar approach has been utilised in other methods \cite{ramos2016hilbert}, \cite{doherty2017bayesian} for map learning.





\section{Results} 
\label{sec:results}
Our method was tested on a simulated indoor data set with a known ground truth, and real data: both the Intel lab data set and Freiburg campus outdoor data set.
\subsection{Simulated results} 
\label{sub:simulated_results}

A simulated indoor environment was used initially to test and compare our proposed method to grid mapping, with a known ground truth. The simulated environment was designed to contain areas where measurements are both dense and sparse, partially occluded areas, and objects smaller than a grid cell. A simulated range sensor was moved around the environment, totalling 24 scans. Each scan consists of 180 range measurements, with a maximum range of 3 metres, at an angular resolution of 2 degrees.

\begin{figure}[h]
\centering
\begin{minipage}{0.49\linewidth}
  \centering
\includegraphics[height=0.9\textwidth]{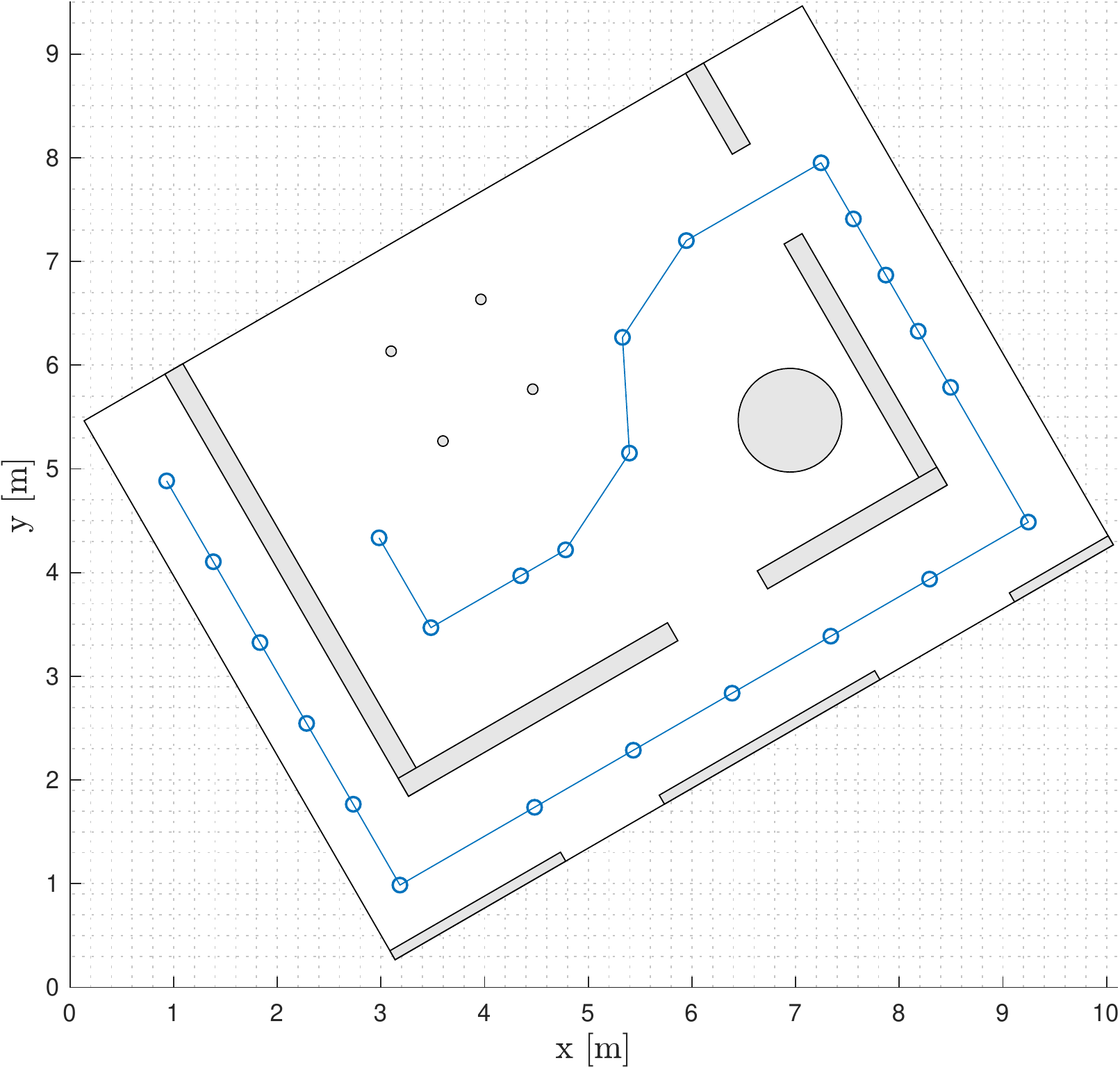}
\subcaption{}\label{fig:GroundTruthSim}
\end{minipage}%
\hfill
\begin{minipage}{0.49\linewidth}
  \centering
\includegraphics[height=0.9\textwidth]{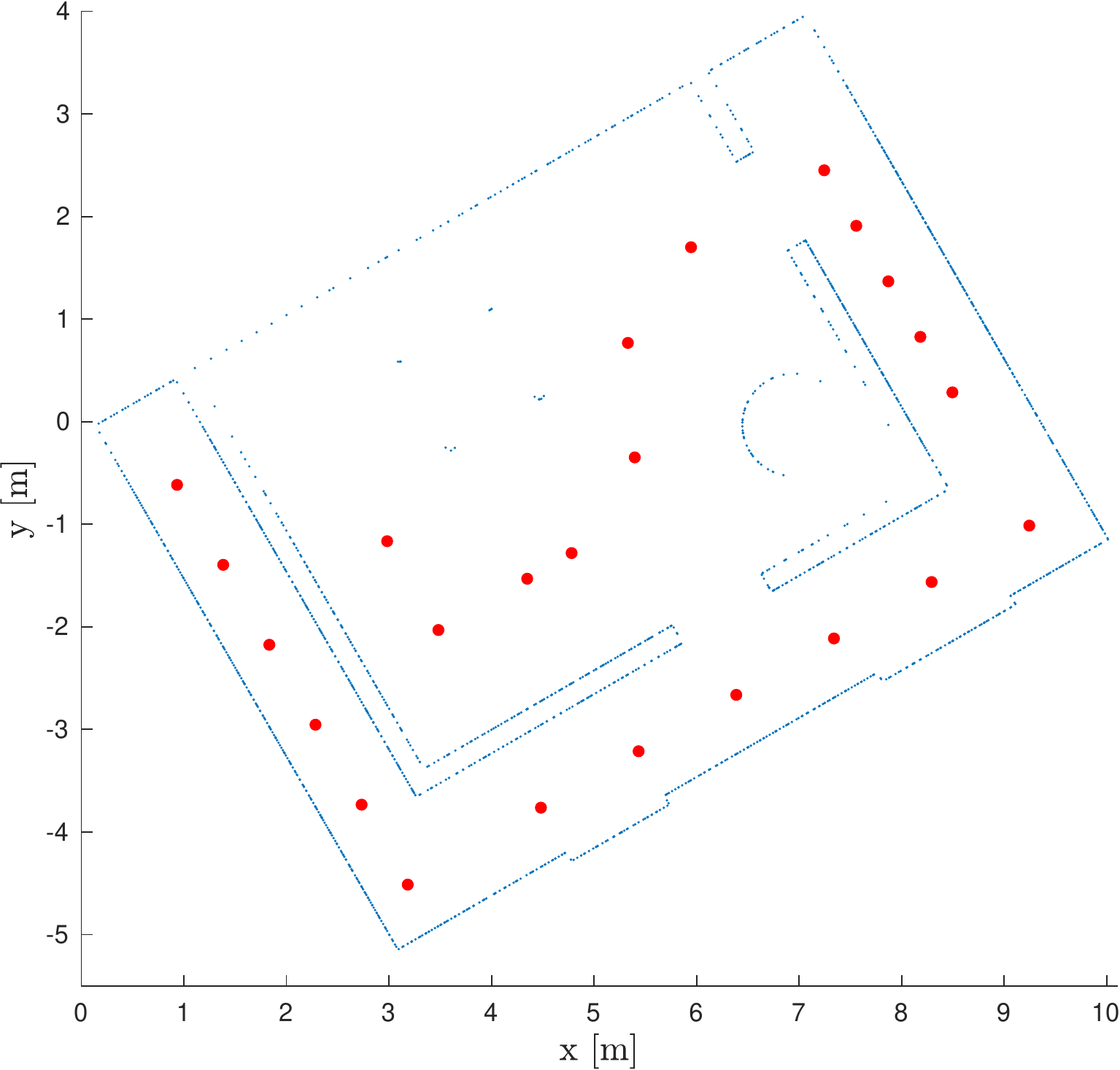}
\subcaption{}\label{fig:measure_sim}
\end{minipage}
\caption{Ground truth (a) and measurement set (b).}
\end{figure}

\begin{figure}[h]
  \vspace{2mm}
\begin{minipage}{1\linewidth}
  \centering
\includegraphics[width=0.99\linewidth]{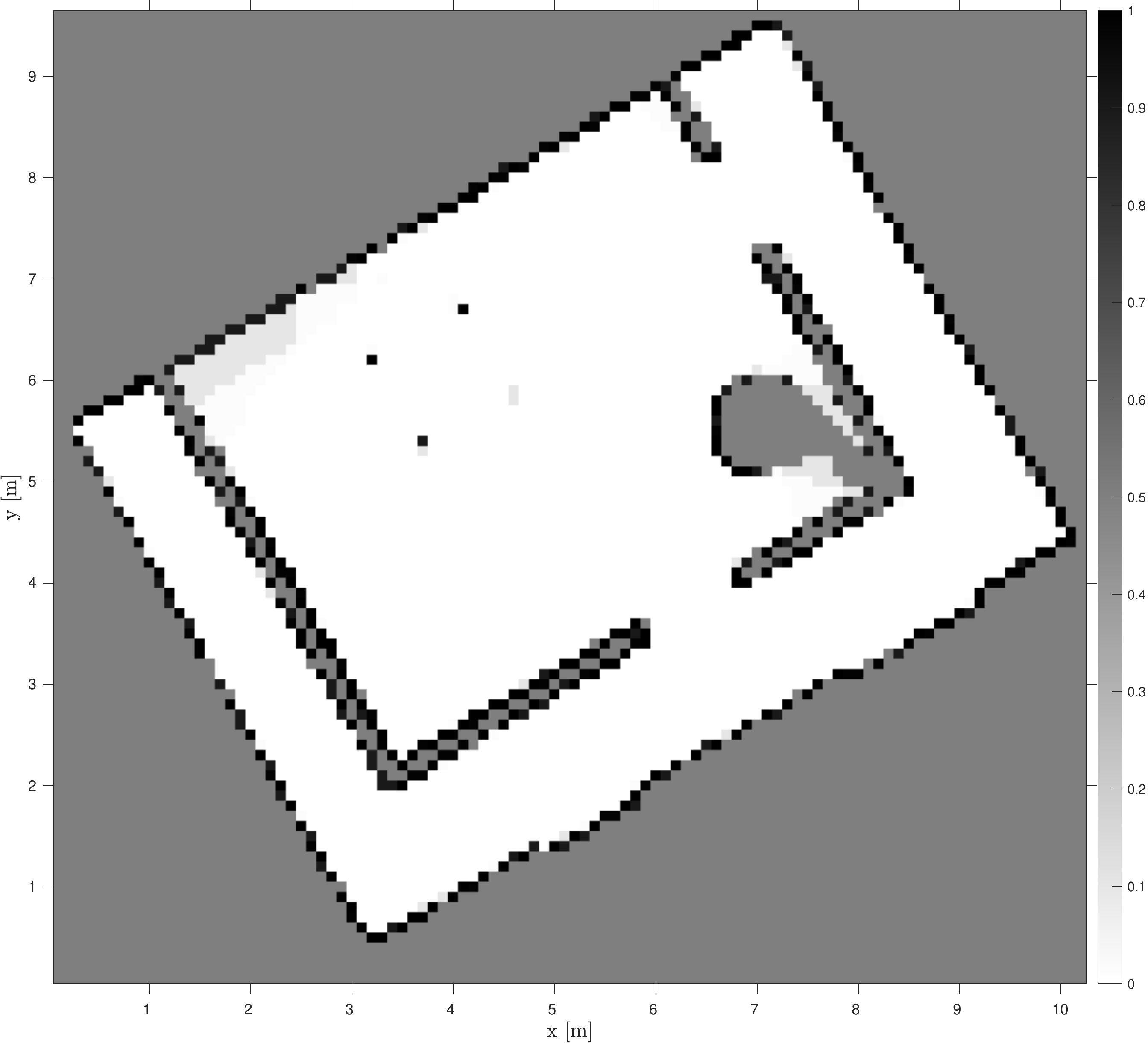}
\subcaption{Grid map.}\label{fig:GridMapSim1e-1}
  \centering
\includegraphics[width=0.99\linewidth]{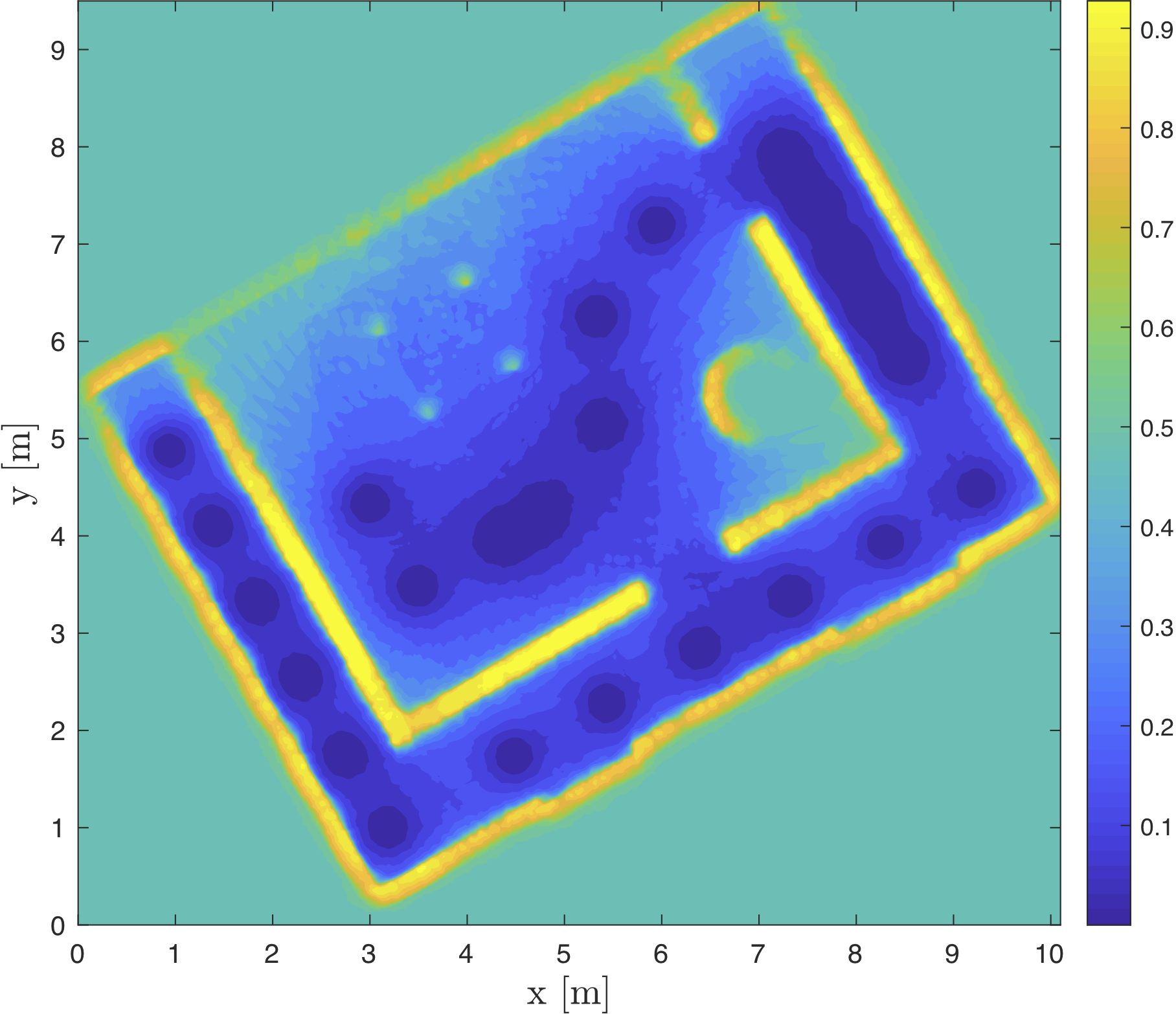}
\subcaption{Proposed method.}\label{fig:IPOM_sim}
\end{minipage}
\caption{Visualisation of a simulated indoor data set. } \label{fig:SimulatedResults}
\end{figure}

The ground truth and robot poses are shown in Fig.~\ref{fig:GroundTruthSim}, with the range measurements shown in Fig.~\ref{fig:measure_sim}. An occupancy grid map with a cell size of 0.1~m was generated for comparative purposes, shown in Fig.~\ref{fig:GridMapSim1e-1}. Fig.~\ref{fig:IPOM_sim} shows our method with the same data set.

\begin{figure}[h]
  \centering
  \includegraphics[width = 0.8\linewidth]{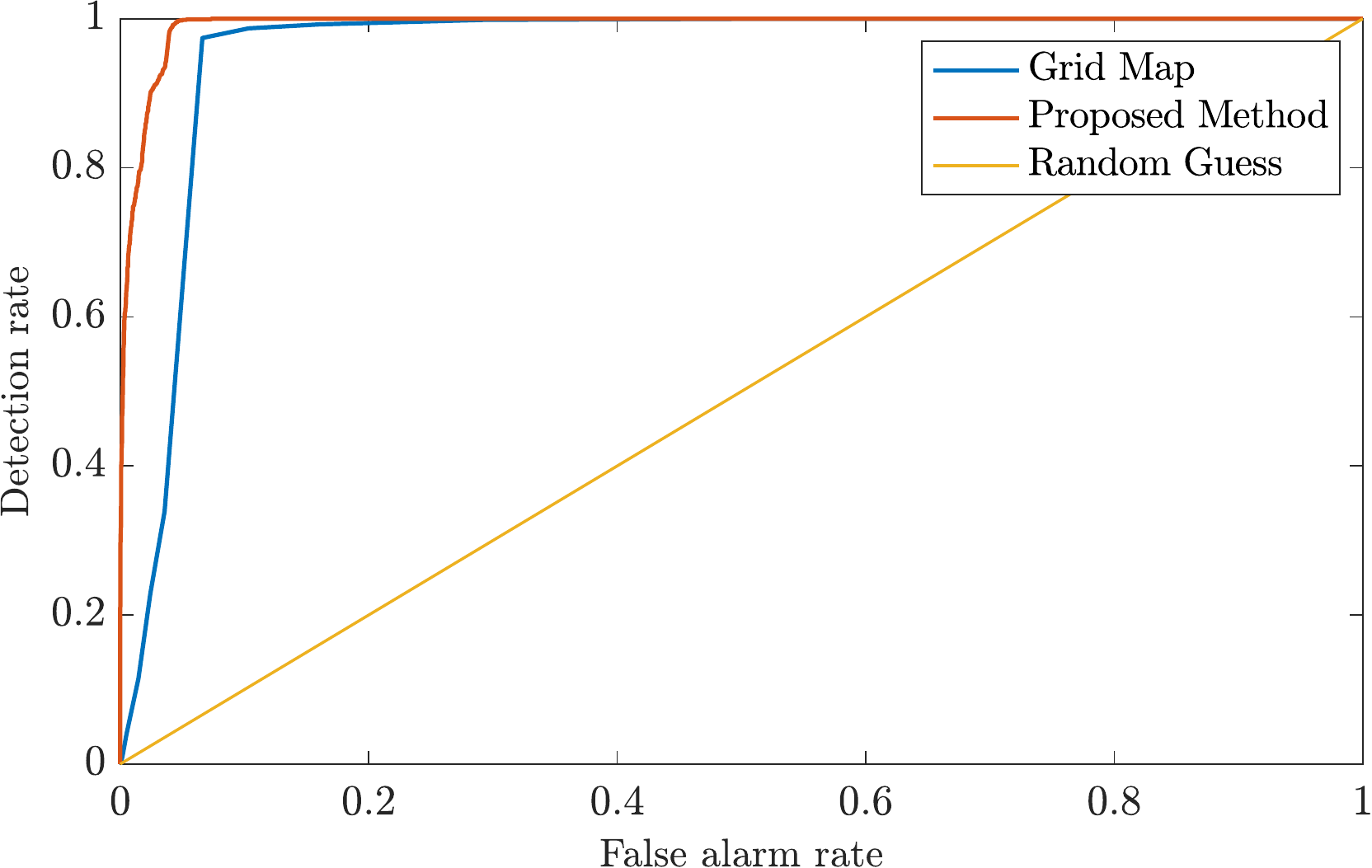}
  \caption{ROC curves for both a grid map and out method created from the simulated indoor environment.}
  \label{fig:ROC}
\end{figure}

A receiver operating characteristic (ROC) curve was generated to compare the accuracy of our method with grid mapping. Fig.~\ref{fig:ROC} depicts detections against false-alarms (false positive), Table.~\ref{tab:roc} shows the false positive rate for both the grid map and our method required to obtain an detection (true positive) rate of $0.95$.

\begin{table}[h]
\vspace{2mm}
  \caption{ROC results.}
  \label{tab:roc}
  \centering
  \begin{tabular}{|l|>{\centering\arraybackslash}m{2.5cm}|>{\centering\arraybackslash}m{2.5cm}|}
  \hline
   & Area under the ROC curve & FP detection rate for TP detection rate of 0.95 \\
  \hline
     Random Guess & 0.5 & 0.95 \\
     Grid map & 0.955 &  0.065 \\
     Proposed Method &  0.992 &  0.038 \\
  \hline
  \end{tabular}
\end{table}

\subsection{Experimental Dataset} 
\label{sub:experimental_dataset}
Two experimental data sets were used to validate our proposed method. Importantly, our method performs well at modelling both small and large scale obstacles in the office environment depicted in the first experiment and the outdoor environment shown in the second. The results show that the method performs well when compared with grid mapping and other continuous occupancy mapping techniques (for a visual comparison see \cite{IntelData}, \cite{FreiburgData} for grid maps generated with the same data set).

\begin{figure}[h]
    \centering
    \includegraphics[width = 0.99\linewidth]{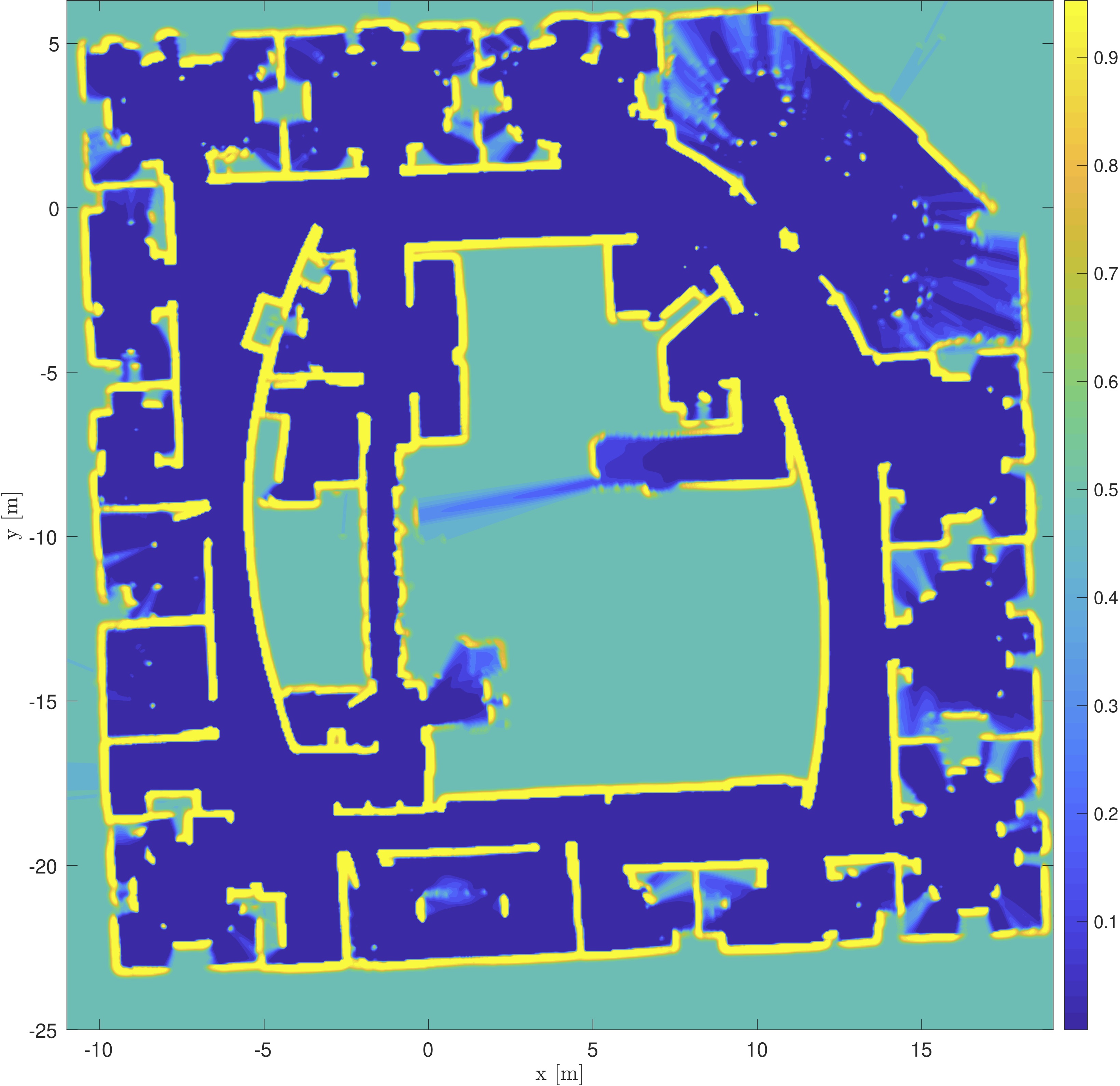}
    \caption{Intel Lab.}
    \label{fig:intel}
  \end{figure}

The first experiment, shown in Fig.~\ref{fig:intel} used the Intel lab data set \cite{IntelData}. Each scan contains 180 range measurements, that covered the $180^\circ$ in front of the robot, with 910 scans in total. The second experiment shown in Fig.~\ref{fig:IPOMFR} uses the Freiburg data set \cite{FreiburgData}. Each scan contains 360 range measurements, that covered the $180^\circ$ in front of the robot, with 2008 scans in total. 

\setlength{\mylengthL}{0.53\textwidth}
\setlength{\mylengthR}{\textwidth - \mylengthL}
\begin{figure*}[h]
\vspace{2mm}
\centering
\begin{minipage}{\mylengthL}
  \centering
\includegraphics[width=0.99\textwidth]{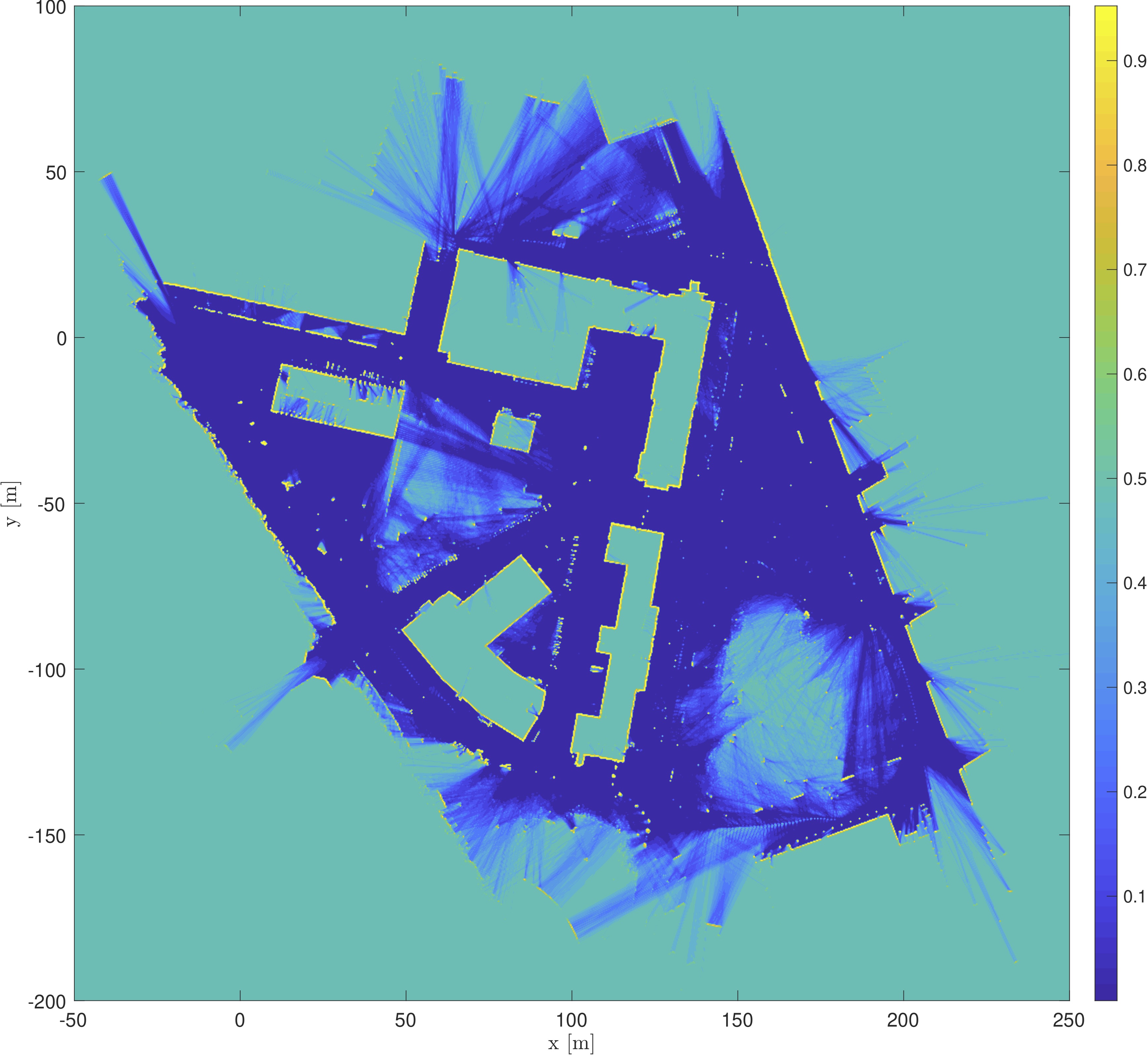}
\subcaption{Freiburg Campus.}\label{fig:frcampus}
\end{minipage}%
\hfill
\begin{minipage}{\mylengthR}
  \centering
\includegraphics[width =0.99\textwidth]{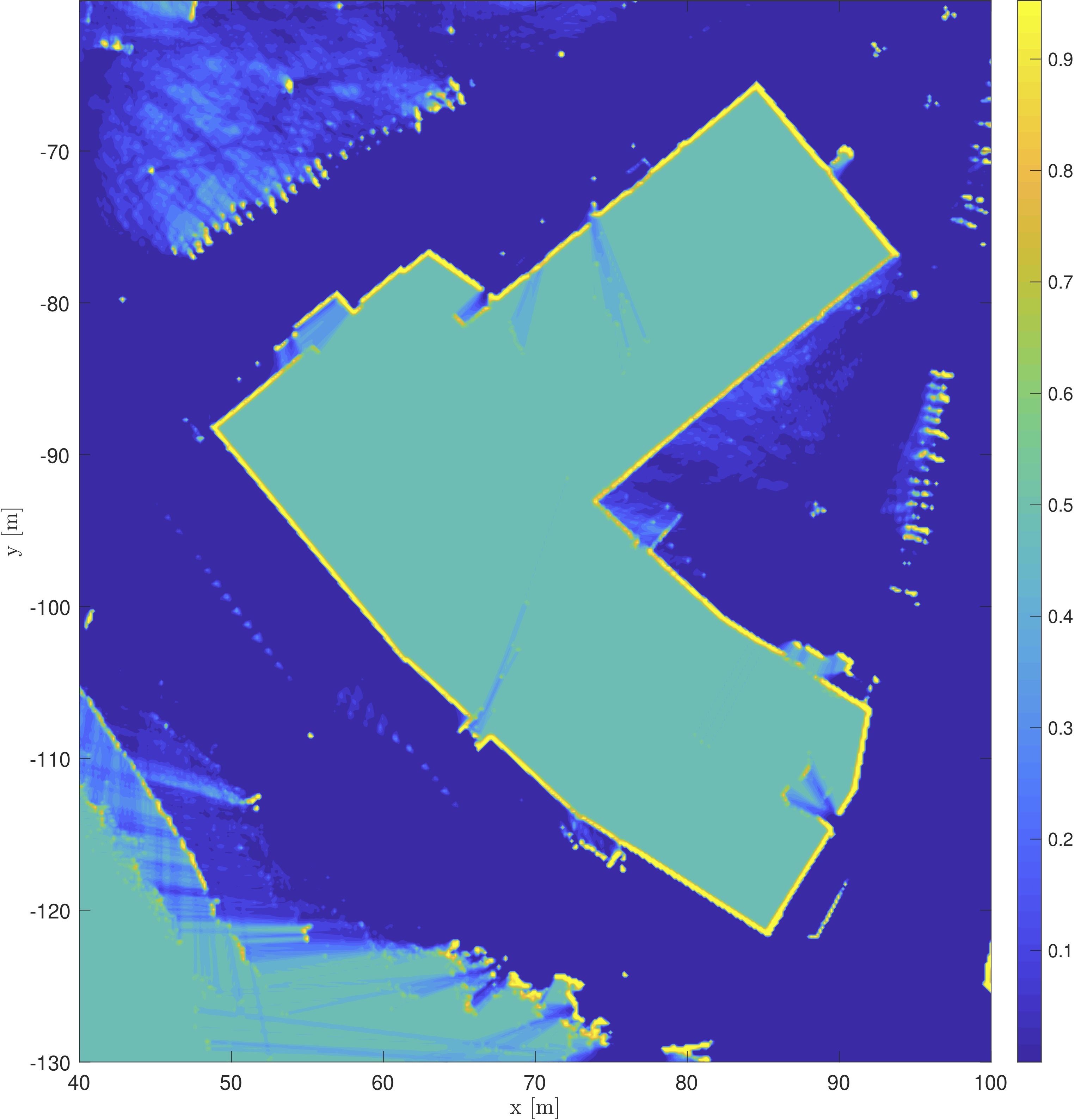}
\subcaption{High resolution Freiburg building.}\label{fig:FRzoom}
\end{minipage}%
\caption{Experimental data sets.} 
\label{fig:IPOMFR}
\end{figure*}

\section{Conclusion} 
\label{sec:conclusion}

In this work we have proposed a new method of learning a continuous occupancy field. This method has several advantages over grid mapping, overcoming geometric limitations and issues of spatial correlation. The accuracy of the proposed method exceeds that of the grid map and is competitive with recently proposed methods. 
The main advantage of our method is that it produces a very accurate map with a linear run time. Our method also does not require a large number of parameters to be trained online for inference, but only a small number of hyperparameters to be trained offline, or before the inference takes place. For this purpose we proposed a modified pseudo likelihood as a cost function for optimising the hyperparameters.


Although the complexity of the algorithm is linear, it requires all measurements to be stored if the probability of occupancy is to be inferred at a new location, which will quickly introduce a storage overhead. In this sense we have not presented a mapping algorithm but formulated  a solution for exact inference. Methods of sparse measurement compression need to be explored to reduce the storage overhead.

\end{document}